%% file: main.tex
\documentclass{article}

\usepackage{microtype}
\usepackage{graphicx}
\usepackage{subcaption}
\usepackage{booktabs} 
\usepackage{stfloats}
\usepackage{hyperref}
\usepackage{booktabs} 
\usepackage[table]{xcolor} 
\newcommand{\reshape}{\mathrm{reshape}}
\usepackage{tabularx}
\usepackage{array}

\usepackage{algpseudocode}

\usepackage{amsmath,amssymb}

\usepackage[preprint]{icml2026}
\makeatletter
\renewcommand{\printAffiliationsAndNotice}[1]{\global\icml@noticeprintedtrue}
\makeatother

\usepackage{xcolor}
\definecolor{hlblue}{HTML}{BFE6FF}
\usepackage{soul}
\sethlcolor{hlblue!35}

\newcommand{\umlaut}[1]{\mathaccent"707F{#1}} 
\newcommand{\Plucker}{Pl\umlaut{u}cker}

\usepackage{amsmath}
\usepackage{amssymb}
\usepackage{mathtools}
\usepackage{amsthm}
\usepackage{multirow}
\usepackage{amssymb}
\usepackage{pifont}
\newcommand{\cmark}{\ding{51}} 
\newcommand{\xmark}{\ding{55}} 

\usepackage[capitalize,noabbrev]{cleveref}

\theoremstyle{plain}

\theoremstyle{definition}

\theoremstyle{remark}

\usepackage[textsize=tiny]{todonotes}

\icmltitlerunning{Predicting Camera Pose from Perspective Descriptions for Spatial Reasoning}

\begin{document}

\twocolumn[
  \icmltitle{Predicting Camera Pose from Perspective Descriptions for Spatial Reasoning}

  \icmlsetsymbol{equal}{*}
    \icmlsetsymbol{inst}{} 

  \begin{icmlauthorlist}
    \icmlauthor{Xuejun Zhang}{inst}
    \icmlauthor{Aditi Tiwari}{inst}
    \icmlauthor{Zhenhailong Wang}{inst}
    \icmlauthor{Heng Ji}{inst}
  \end{icmlauthorlist}

  \vspace{0.5em}
{\centering\small
University of Illinois Urbana-Champaign\\
\vspace{0.25em}
\texttt{xuejunz2@illinois.edu,\; hengji@illinois.edu}
\par}
  \vspace{0.6em}

  \icmlkeywords{Machine Learning, ICML}

  \vskip 0.3in
]

\printAffiliationsAndNotice{}  

\input{sections/00-abstract}
\input{sections/01-introduction}
\input{sections/02-background}

\input{sections/03-method}
\input{sections/04-experiment}
\input{sections/05-discussion}
\input{sections/06-conclusion}
\input{sections/07-acknowledgementy}

\newpage

\bibliography{example_paper}
\bibliographystyle{icml2026}

\input{sections/a_appendix}

\end{document}

%% file: sections/00-abstract.tex
\begin{abstract}

Multi-image spatial reasoning remains challenging for current multimodal large language models (MLLMs). While single-view perception is inherently 2D, reasoning over multiple views requires building a coherent scene understanding across viewpoints. In particular, we study perspective taking, where a model must build a coherent 3D understanding from multi-view observations and use it to reason from a new, language-specified viewpoint. We introduce \textsc{CamCue}, a pose-aware multi-image framework that uses camera pose as an explicit geometric anchor for cross-view fusion and novel-view reasoning. \textsc{CamCue} injects per-view pose into visual tokens, grounds natural-language viewpoint descriptions to a target camera pose, and synthesizes a pose-conditioned imagined target view to support answering. To support this setting, we curate \textsc{CamCue-Data} with 27{,}668 training and 508 test instances pairing multi-view images and poses with diverse target-viewpoint descriptions and perspective-shift questions. We also include human annotated viewpoint descriptions in the test split to evaluate generalization to human language. \textsc{CamCue} improves overall accuracy by 9.06\% and predicts target poses from natural-language viewpoint descriptions with over 90\% rotation accuracy within 20$^\circ$ and translation accuracy within a 0.5 error threshold. This direct grounding avoids expensive test-time search-and-match, reducing inference time from 256.6s to 1.45s per example and enabling fast, interactive use in real-world scenarios.
\noindent Project page: \url{https://xuejunzhang2002.github.io/camcue/}

\end{abstract}

%% file: sections/01-introduction.tex
\section{Introduction}
\input{floating/teaser}
Spatial intelligence moves beyond single-image perception and naive multi-image aggregation. Rather than treating each view as an independent 2D snapshot, an agent needs to connect views via their spatial relationships to form a coherent 3D understanding that supports reasoning beyond the observed images~\citep{chen2024and, gholami2025spatial, wang20253d, yin2025spatial, zhao2025spacemind, lee2025perspective, yeh2025seeinganotherperspectiveevaluating, yang2025mmsi}. Humans do this naturally: when told “sit on the sofa behind the black table,” we can mentally relocate to that viewpoint and imagine what we would see, then answer questions from that perspective~\citep{article, Meilinger2011}, which is illustrated by Figure~\ref{fig:teaser}. However, current multimodal large language models (MLLMs) still struggle with this kind of perspective taking. Even with multiple context images, they often fail to reliably ground a language-specified viewpoint and reason from the intended perspective ~\citep{lee2025perspective, yeh2025seeinganotherperspectiveevaluating, xu2025spatialbench, yin2025spatial}. This gap motivates our study of language-guided viewpoint grounding for multi-view spatial reasoning. We study perspective-shift reasoning where the target viewpoint is specified in natural language. Given multiple context images and a question, the model needs to ground the description to a target camera pose and answer from that perspective.

A recent line of work tackles perspective-shift reasoning by augmenting MLLMs with generative world models that actively synthesize additional observations at inference time~\citep{lee2025perspective, yang2025mindjourney}. While promising, existing pipelines are often built around a single reference view and do not effectively integrate multiple contextual images as a unified source of evidence~\citep{lee2025perspective, yang2025mindjourney}. In addition, most controllable generators are largely query-agnostic, which can produce imagined views that are irrelevant or even inconsistent with the downstream question~\citep{yang2025mindjourney}. Many of these methods rely on expensive test-time procedures such as iterative search or multiple candidate rollouts to obtain a useful imagined view, resulting in high latency and limited practicality. Finally, off-the-shelf novel-view synthesis is typically pose-conditioned, whereas MLLMs do not reliably infer target camera poses from natural language description, leaving a mismatch between language-driven viewpoint specification and pose-controlled generation~\citep{jin2024lvsm, zhou2025stable}.

To address these limitations, we introduce \textsc{CamCue}, a pose-aware multi-image MLLM framework that can predict the camera pose of the language-specified target perspective. Camera pose provides a compact, explicit representation of viewpoint that situates each image in a shared 3D coordinate frame, making inter-view geometry directly usable for multimodal reasoning~\citep{liao2025thinking, zhao2025spacemind}. Our key design principle is to make viewpoint an explicit geometric anchor for multi-view reasoning. We start by injecting per-view camera information into the corresponding visual features, so that the model can align evidence across images through geometry rather than treating each image as individual input. We then interpret the natural-language target-perspective description by mapping it to a concrete target camera pose, which specifies where the model should ``mentally stand'' to answer the question. Conditioned on this predicted pose, we further synthesize the corresponding target-view image and treat the imagined observation as additional evidence for answering. This tight coupling between language-specified perspective, pose prediction, and pose-conditioned view synthesis strengthens multi-image fusion and substantially improves performance on perspective-shift spatial reasoning.

To support this setting, we curate \textsc{CamCue-Data}, a dataset tailored to perspective-shift reasoning. 
\textsc{CamCue-Data} contains 27{,}668 training instances and 508 test instances, and pairs multi-view images and per-view camera poses with diverse natural-language target-perspective descriptions, including human-annotated descriptions, and questions that require answering from the specified viewpoint. 
On this benchmark, \textsc{CamCue} yields substantial gains on perspective-shift spatial reasoning, improving overall accuracy by 9.06\%. 
It also predicts target camera poses directly from natural-language descriptions with strong accuracy, achieving over 90\% rotation accuracy within 20$^\circ$ and translation accuracy at t@0.5. 
Moreover, by predicting an explicit target pose, \textsc{CamCue} avoids expensive test-time search-and-match used by prior methods, reducing inference time from 256.6 seconds to 1.45 seconds per example and enabling fast, interactive use in real-world scenarios.
Beyond \textsc{CamCue-Data}, \textsc{CamCue} also improves performance on general multi-image spatial reasoning benchmarks such as MindCube Tiny~\citep{yin2025spatial} and MMSI~\citep{yang2025mmsi}.
                         
In summary, our contributions are listed as follows:
\begin{itemize}
    \item We propose \textsc{CamCue}, a pose-aware multi-image MLLM framework that injects per-view camera information into the corresponding visual features, enabling geometry-aware fusion across views for spatial reasoning.

    \item \textsc{CamCue} can map a natural-language target-view description to an explicit target camera pose, providing a concrete viewpoint representation for answering from the specified perspective.

    \item Conditioned on the predicted target pose, \textsc{CamCue} synthesizes the corresponding target-view image and feeds it back as additional evidence, substantially improving perspective-shift reasoning ability.

    \item We curate \textsc{CamCue-Data}, a dataset tailored to perspective-shift reasoning that pairs multi-view images with diverse, detailed natural-language camera viewpoint descriptions, including human-annotated descriptions, and target-view questions that require reasoning from the described viewpoint.
\end{itemize}

%% file: floating/teaser.tex
\begin{figure*}[!t]
    \centering
    \includegraphics[width=0.9\textwidth]{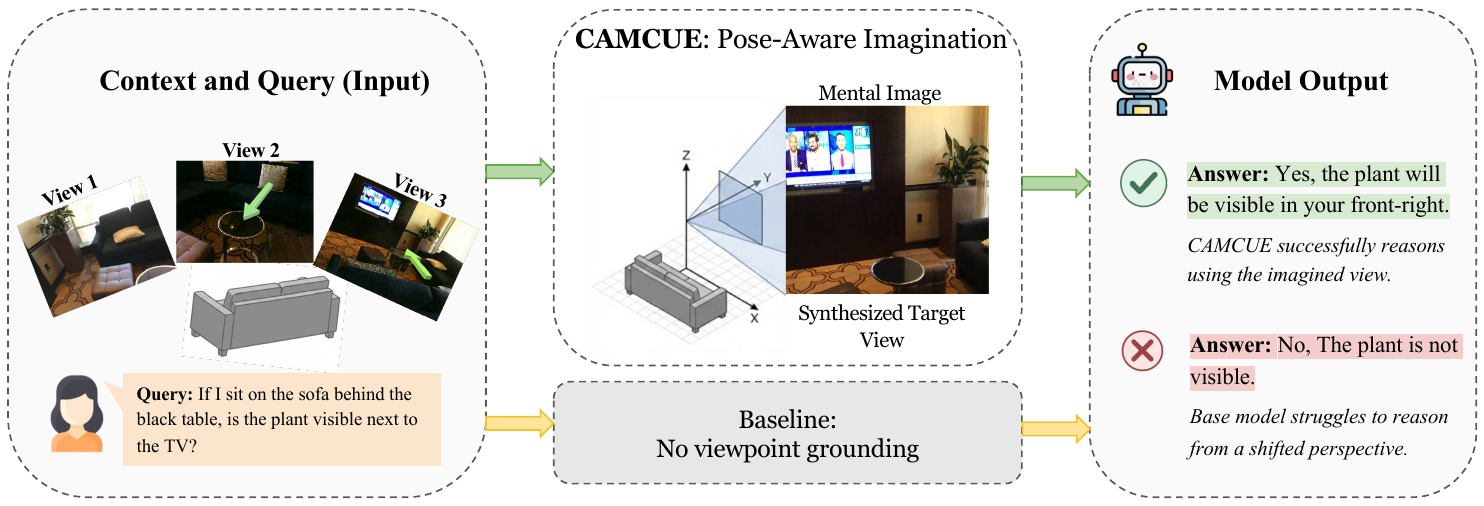}
    \caption{Perspective-shift reasoning with CamCue. Given multi-view context images, CamCue maps a natural-language viewpoint description to an explicit target camera pose and synthesizes the corresponding target view for reliable spatial reasoning.} 
    \vspace{-10pt}
    \label{fig:teaser}
\end{figure*}

%% file: sections/02-background.tex
\section{Related Work}

\subsection{Multi-Image Spatial Reasoning Benchmarks}

Multi-image spatial reasoning is a key probe for evaluating spatial intelligence in MLLMs, as it requires integrating partial observations from multiple viewpoints into a coherent and viewpoint-consistent scene understanding. Recent benchmarks reveal substantial gaps between current MLLMs and human performance, with models often struggling to fuse evidence across views and maintain consistent spatial beliefs. Representative datasets include MindCube~\citep{yin2025spatial}, SpatialBench~\citep{xu2025spatialbench}, MMSI-Bench~\citep{yang2025mmsi}, All-Angles Bench~\citep{yeh2025seeinganotherperspectiveevaluating}, and ViewSpatial-Bench~\citep{li2025viewspatialbenchevaluatingmultiperspectivespatial}. Surveys and diagnostic studies further organize these benchmarks by cognitive demands and emphasize that reliable multi-view integration remains challenging~\citep{liu2025spatialreasoningmultimodallarge,zhang2025mllmsstrugglespatialunderstanding,yu2025farvlmsvisualspatial}. These findings motivate methods that explicitly ground viewpoints and align observations across views, such as pose-aware approaches that use camera pose as a geometric anchor for multi-view fusion and perspective-consistent reasoning~\citep{chen2024and, liao2025thinking}.

\subsection{Perspective-Taking and Allocentric Reasoning in MLLMs}

Beyond reasoning within a single image, many embodied and multi-view tasks require perspective taking, where the model answers questions from an alternative viewpoint that is unobserved and specified in natural language. This setting calls for allocentric scene understanding and reliable viewpoint grounding, yet current MLLMs can be brittle under perspective shifts even with multiple context images~\citep{sqa3d,yin2025spatial,yang2025mmsi,yeh2025seeinganotherperspectiveevaluating,li2025viewspatialbenchevaluatingmultiperspectivespatial}. Recent approaches attempt to bridge this gap via mental imagery or generative rollouts that synthesize missing observations at inference time~\citep{lee2025perspective,yang2025mindjourney,cao2025spatialdreamer}. While effective in some cases, these pipelines typically do not explicitly ground the language-specified viewpoint to a concrete target pose, and instead rely on searching over candidate motions and viewpoints. As a result, synthesized observations may drift from the intended viewpoint described in language, producing evidence that is misaligned with the target perspective. Moreover, searching over many candidates can be computationally expensive, making such approaches less suitable for applications that require timely, interactive feedback.

\subsection{Language-Grounded Viewpoint Imagination}
A key challenge underlying perspective taking is to imagine a faithful observation from a language-specified, unobserved viewpoint and use it to support cross-view alignment and spatial reasoning. Existing approaches largely fall into two lines. One line relies on pose-free image generation and editing models to synthesize a new view directly from text and contextual images~\citep{team2023gemini, wu2025qwen, achiam2023gpt}. However, such generations provide no explicit control over camera information and may not be reliably 3D-consistent under perspective shifts, making the imagined evidence brittle for viewpoint-sensitive reasoning. The other line uses pose-conditioned novel-view synthesis models, which can produce geometrically consistent renderings when a target pose is given ~\citep{zhou2025stable, jin2024lvsm, yu2021pixelnerf, charatan2024pixelsplat, chen2024mvsplat, wu2024reconfusion, gao2024cat3d}, but they do not address the crucial missing step in our setting: mapping a natural-language viewpoint description to the target camera pose. \textsc{CamCue} bridges these two lines by learning to predict the target camera pose from language and using it as an explicit geometric anchor for token-level fusion and image imagination.

%% file: sections/03-method.tex
\section{Method}
\label{sec:method}
\input{floating/framework}
In this section, we introduce \textsc{CamCue}, a pose-aware multi-image framework for perspective-shift spatial reasoning.
Figure~\ref{fig:framework} provides an overview of the \textsc{CamCue} pipeline.
Given a text prompt $T$ that contains a natural-language description of a target perspective and a question, together with a set of $V$ contextual images $\mathcal{I}=\{I_i\}_{i=1}^{V}$ and their associated camera poses $\mathcal{P}=\{P_i\}_{i=1}^{V}$, \textsc{CamCue} predicts the answer under the specified target perspective.
We first present the \textsc{CamCue} model architecture in Sec.~\ref{sec:method_arch}. 
We then describe the construction of \textsc{CamCue-Data} in Sec.~\ref{sec:data}.

\subsection{Architecture}
\label{sec:method_arch}

\paragraph{Pl\"ucker encoder}

As shown in Fig.~\ref{fig:framework}, each contextual view provides camera extrinsics $\mathbf{C}_i \in \mathbb{R}^{4\times 4}$ and intrinsics $\mathbf{K}_i$. 

Following prior work~\citep{jiang2025rayzer}, we transform $(\mathbf{C}_i,\mathbf{K}_i)$ into a pixel-aligned Pl\"ucker ray map
\begin{equation}
\mathbf{R}_i = \Plucker (\mathbf{C}_i,\mathbf{K}_i)\in\mathbb{R}^{H\times W\times 6},
\end{equation}
which represents the camera pose information as dense rays aligned with image pixels.

We then encode $\mathbf{R}_i$ into patch-aligned camera tokens
\begin{equation}
Z_i = E_{\text{pose}}(\mathbf{R}_i)\in\mathbb{R}^{S\times d},
\end{equation}
where $S = H_p W_p$ is the number of patch tokens under the backbone's canonical resolution, with $H_p = H'/p$ and $W_p = W'/p$ for patch size $p$.

$E_{\text{pose}}$ is a lightweight Pl\"ucker encoder that follows the same patchification and spatial aggregation scheme as the vision backbone.
Specifically, $E_{\text{pose}}$ follows the same tokenization pipeline as the vision encoder: it first resizes $\mathbf{R}_i$ to the backbone's canonical resolution, then patchifies it and applies a patch embedding to convert each local ray patch into a $d$-dimensional token. This yields a patch-aligned token grid that is spatially aligned with the image patch tokens for subsequent fusion.

\paragraph{Pose-aware token fusion.}
Given the image patch tokens $X_i \in \mathbb{R}^{S\times d}$ from the vision backbone and the corresponding Pl\"ucker camera tokens $Z_i \in \mathbb{R}^{S\times d}$, we fuse pose information into the visual representation in a patch-aligned manner. We concatenate tokens at the same patch index and apply a lightweight MLP projection to produce a residual update:
\begin{equation}
\tilde{X}_i = X_i + W \,[Z_i; X_i],
\end{equation}
where $[\cdot;\cdot]$ denotes feature-wise concatenation and $W\in\mathbb{R}^{d\times 2d}$.
This design preserves the backbone token layout while injecting per-patch geometric cues, yielding fused tokens $\tilde{X}_i \in \mathbb{R}^{S\times d}$ for subsequent multi-view reasoning.

\paragraph{Target pose prediction.}

As shown in Fig.~\ref{fig:framework} (the \textit{Pose Adapter} branch), given the fused multi-view scene tokens $\tilde{X} \in \mathbb{R}^{T_{\text{vis}}\times d}$ and the text hidden states $H \in \mathbb{R}^{T_{\text{text}}\times d}$, we predict the target camera pose with a query-based cross-attention head.
Concretely, we introduce $N$ learnable query vectors $Q_0 \in \mathbb{R}^{N\times d}$ that attend to the concatenated sequence of text and visual tokens:
\begin{equation}
Y = \mathrm{Attn}\!\left(Q_0,\,[H;\tilde{X}],\,[H;\tilde{X}]\right)\in\mathbb{R}^{N\times d},
\end{equation}
where $\mathrm{Attn}(\cdot)$ is multi-head attention and $[\cdot;\cdot]$ denotes concatenation along the token dimension.
We then project each attended query with a linear projection to obtain pose query tokens
\begin{equation}
U = \psi(Y)\in\mathbb{R}^{N\times d_q}.
\end{equation}
We set $N=16$ and map the $16$ pose query tokens to a camera-to-world matrix:
\begin{equation}
\hat{\mathbf{C}}_{\text{tgt}}=\reshape\!\left(g(U)\right)\in\mathbb{R}^{4\times 4},
\end{equation}
where $g(\cdot)$ produces one scalar per token.

\paragraph{Answer generation.}
Our model produces the language answer and the target pose prediction in a single pass.
Concretely, the model autoregressively generates a response sequence that begins with a pose slot segment and then outputs the final text answer.
At inference time, we optionally use the predicted camera pose to synthesize an imagined target observation with an image decoder; in our implementation, we use LVSM ~\citep{jin2024lvsm}.
The synthesized image is then treated as additional visual evidence and provided to the MLLM to answer the question again, yielding an evidence-enhanced prediction.

\paragraph{Training objective.}
We train the model with a weighted sum of language modeling and pose regression losses:
\begin{equation}
\mathcal{L}=\lambda_{\text{lang}}\mathcal{L}_{\text{lang}}+\lambda_{\text{pose}}\mathcal{L}_{\text{pose}}.
\end{equation}
$\mathcal{L}_{\text{lang}}$ is the standard cross-entropy loss on text output. 
$\mathcal{L}_{\text{pose}}$ supervises the predicted target-view camera extrinsics $\hat{\mathbf{C}}_{\text{tgt}}$ with the ground-truth extrinsics $\mathbf{C}_{\text{tgt}}$:
\begin{equation}
\mathcal{L}_{\text{pose}}
=
\mathrm{MSE}(\hat{\mathbf t}, \mathbf t)
+
\mathrm{MSE}(\hat{\mathbf R}, \mathbf R),
\end{equation}
where $(\mathbf{R},\mathbf{t})$ and $(\hat{\mathbf{R}},\hat{\mathbf{t}})$ denote the rotation and translation components extracted from $\mathbf{C}_{\text{tgt}}$ and $\hat{\mathbf{C}}_{\text{tgt}}$, respectively, i.e., $\mathbf{R}=(\mathbf{C}_{\mathrm{tgt}})_{1:3,1:3}$, $\mathbf{t}=(\mathbf{C}_{\mathrm{tgt}})_{1:3,4}$ and similarly $\hat{\mathbf{R}}=(\hat{\mathbf{C}}_{\mathrm{tgt}})_{1:3,1:3}$, $\hat{\mathbf{t}}=(\hat{\mathbf{C}}_{\mathrm{tgt}})_{1:3,4}$.
\input{floating/dataset_comparison}

\input{floating/data_distribution}

\subsection{Data Construction}
\label{sec:data}

We construct \textsc{CamCue-Data} to evaluate and train perspective-shift spatial reasoning under a realistic multi-view setting, where models must answer questions from a new camera pose position described in language while grounding on a sparse set of contextual observations. As summarized in Table~\ref{tab:data_compare}, existing resources typically cover only a subset of these requirements. 
Many prior situated question answering datasets assume access to a complete 3D scene representation, rather than sparse multi-view observations. 
Meanwhile, recent multi-image benchmarks may include viewpoint language as cues embedded in questions, but often do not provide an explicit, detailed target-view pose description paired with camera pose information. 

To fill this gap, we curate \textsc{CamCue-Data} with 27{,}668 training QA pairs and 508 test QA pairs. Each example pairs sparse multi-view observations and their camera poses with a diverse, detailed natural-language description of a novel target viewpoint. The test set further includes expert-annotated viewpoint descriptions, allowing us to assess robustness and performance in realistic interactive scenarios. Example data samples are provided in Appendix~\ref{app:data_samples}.

\vspace{-10pt}

\paragraph{Data Collection \& Preprocessing.}
We derive training and test QA pairs from the ScanNet training and test splits, respectively. For each scene, we form a multi-view sample by choosing one target view and selecting four contextual views. We first filter candidate contextual views by a moderate translation range to the target and require sufficient viewpoint change so that contextual views are not near-duplicates. We then select four contextual views to ensure the target view is well supported by the contextual observations, using depth-based visibility checks to verify that the contextual views jointly cover what would be seen from the target viewpoint. We keep only target-context groups that pass this visibility criterion and discard redundant groups that are too similar in pose. Qualitative examples are shown in Figure~\ref{fig:render_example}, and details are deferred to the Appendix~\ref{app:view_selection}.

\vspace{-10pt}
\paragraph{Target Pose Descriptions.}
Each data sample includes a target pose description that specifies the novel camera location and orientation. We generate these descriptions with GPT-4.1~\citep{achiam2023gpt} and diversify the phrasing to improve robustness. We use three description styles: layout-anchored descriptions place the camera with respect to the overall room layout; landmark-relative descriptions specify the viewpoint via relative relations among objects; and object-centric descriptions center the viewpoint around a dominant furniture landmark. To evaluate whether models generalize to human-written pose descriptions, we additionally collect expert-written descriptions for the test split. Annotators are instructed to describe the camera position and viewing direction of the target image in clear natural English and to avoid ambiguous phrasing or references so that the described viewpoint is uniquely identifiable.

\vspace{-10pt}
\paragraph{Question Construction.}
Given the contextual views, target pose description, and target view, we curate QA pairs that require answering from the described target perspective. Following prior spatial reasoning benchmarks, we organize questions into five types: \textit{Attribute}, \textit{Count}, \textit{Distance Order}, \textit{Relative Relation}, and \textit{Visibility}. Attribute queries an object's attribute; Count queries the number of instances; Distance Order compares which object is closer to the camera; Relative Relation queries relative spatial relations between objects; Visibility queries whether an object is visible from the target viewpoint; see Appendix~\ref{app:data_samples} for concrete examples. The distribution across training and test splits is shown in Figure~\ref{fig:qa_type_dist}. We generate QA pairs with GPT-4.1, and manually review the test split to ensure questions are unambiguous and answers are correct.

%% file: floating/framework.tex
\begin{figure*}[!t]
    \centering
    \includegraphics[width=0.9\textwidth]{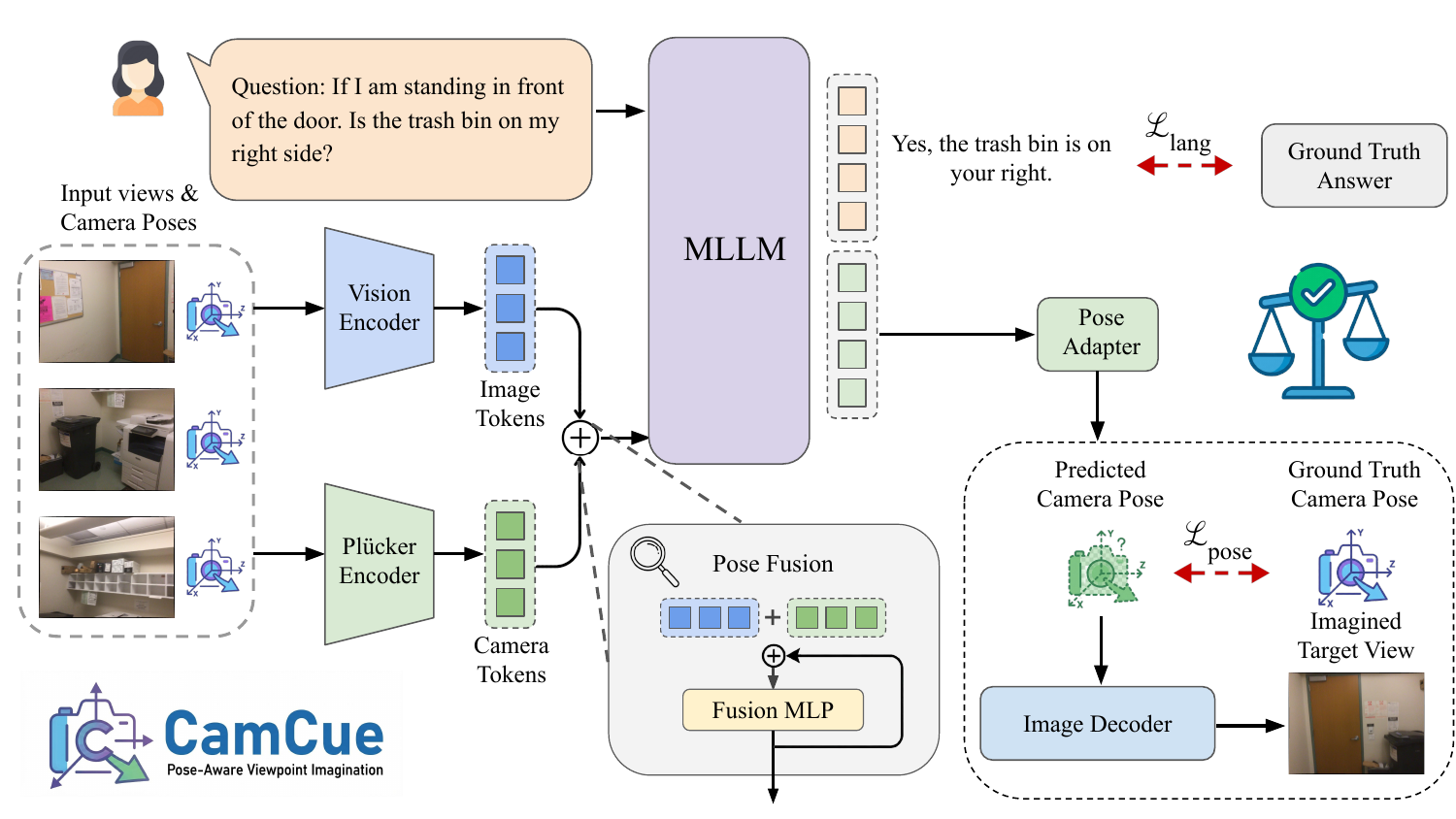}

    \caption{Given multiple contextual images with their camera poses and a natural-language target-viewpoint description plus question, CamCue encodes visual content and pixel-aligned camera pose features, fuses them into pose-aware visual tokens, and uses an MLLM with a pose adapter to jointly generate the answer and predict the target camera pose. The predicted pose can further condition an image decoder to synthesize an imagined target view, which is fed back as additional evidence for answering.}
    \label{fig:framework}
\end{figure*}

%% file: floating/dataset_comparison.tex
\begin{table}[t]
\centering
\small
\setlength{\tabcolsep}{4.0pt}
\renewcommand{\arraystretch}{1.05}
\caption{
Comparison between \textsc{CamCue} and related datasets.
\textit{Camera Pose} is marked when per-view pose metadata is included in the released benchmark. \textit{Pose Desc.} is marked when the benchmark includes perspective taking data, and \textit{Target View} denotes the corresponding ground-truth image.
}
\label{tab:data_compare}

\resizebox{\columnwidth}{!}{%
\begin{tabular}{lcccc}
\toprule
\textbf{Dataset} &
\textbf{Camera Pose} &
\textbf{Multi-img} &
\textbf{Pose Desc.} &
\textbf{Target View} \\
\midrule
ScanQA~\citep{azuma2022scanqa}      & \cmark & \xmark & \xmark & \xmark \\
SQA3D~\citep{sqa3d}        & \cmark & \xmark & \cmark & \xmark \\
MMSI~\citep{yang2025mmsi}          & \xmark & \cmark & \cmark & \xmark \\
MindCube~\citep{yin2025spatial}  & \xmark & \cmark & \cmark & \xmark \\
VSI-Bench~\citep{yang2025thinking} & \xmark & \xmark & \cmark & \xmark \\
All-Angles~\citep{yeh2025seeinganotherperspectiveevaluating} & \xmark & \cmark & \cmark & \xmark \\
SAT~\citep{ray2024sat}            & \xmark & \cmark\textit{(1/2)} & \cmark & \xmark \\
\midrule
\textsc{CamCue} (Ours)     & \cmark & \cmark & \cmark & \cmark \\
\bottomrule
\end{tabular}%
}
\end{table}

%% file: floating/data_distribution.tex
\begin{figure}[t]
\centering
\begin{subfigure}[t]{0.49\columnwidth}
  \centering
  \includegraphics[width=\linewidth]{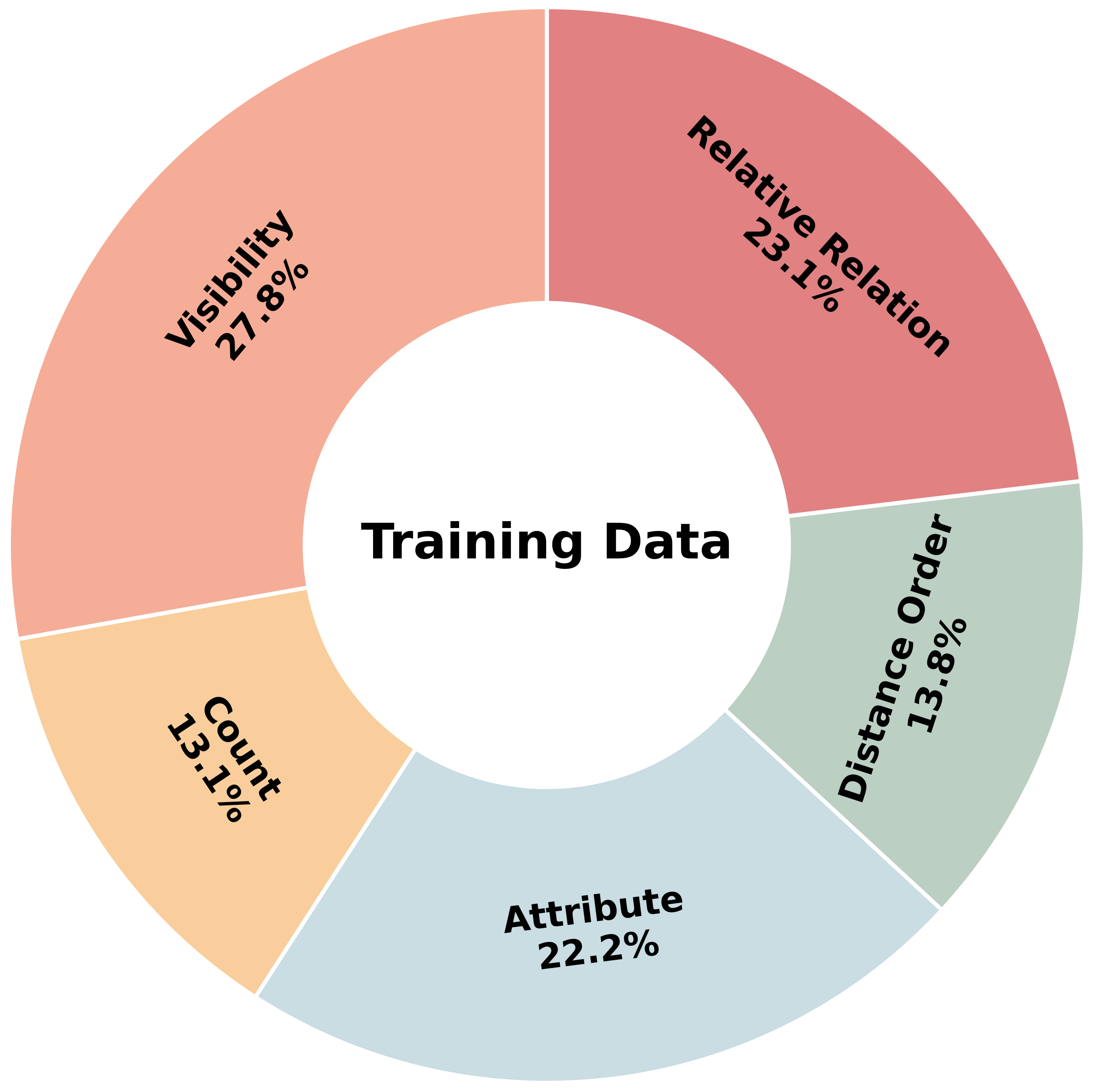}
  \caption{Training data}
\end{subfigure}\hfill
\begin{subfigure}[t]{0.49\columnwidth}
  \centering
  \includegraphics[width=\linewidth]{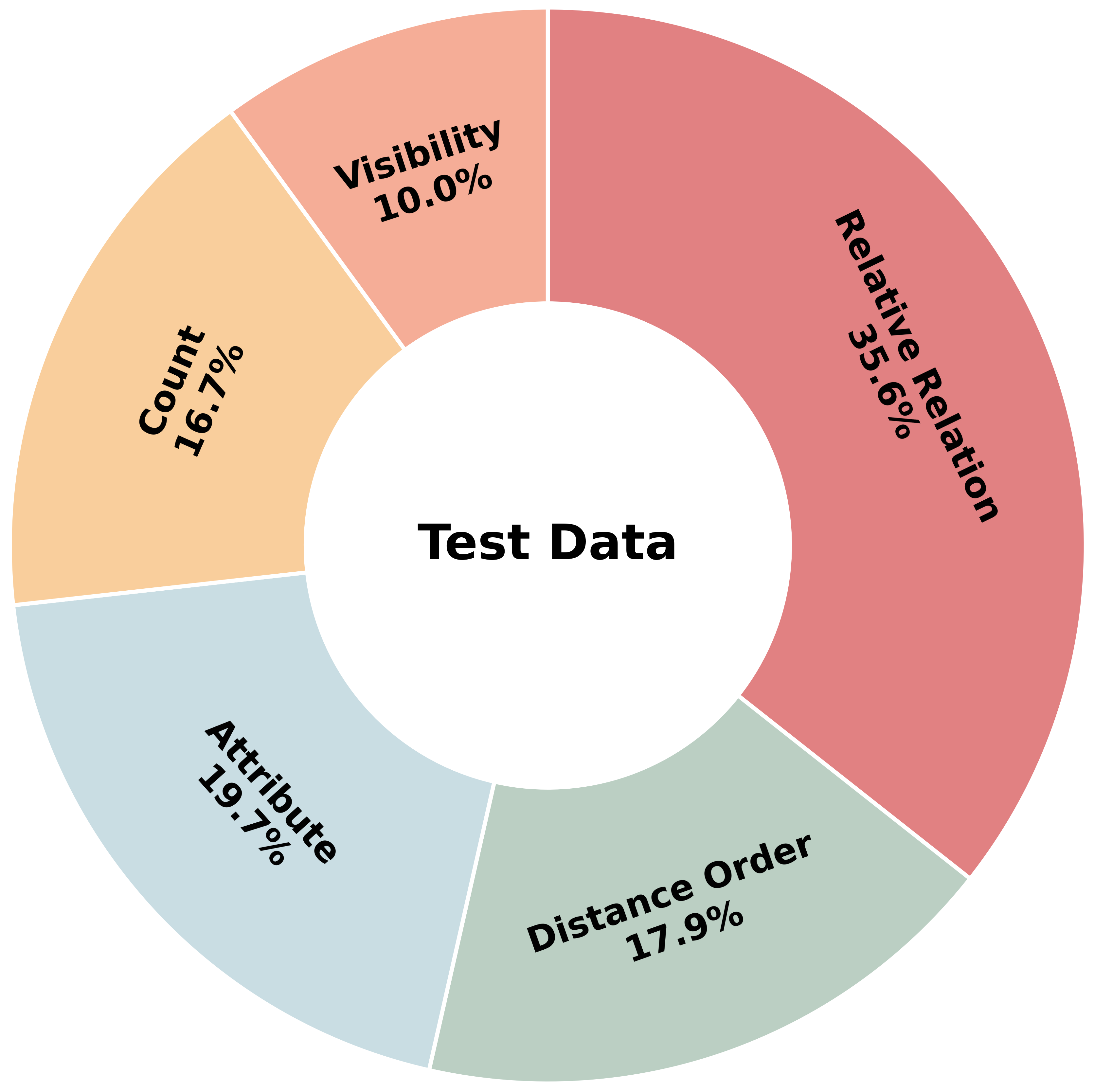}
  \caption{Test data}
\end{subfigure}
\caption{QA type distribution in training and test splits.}
\label{fig:qa_type_dist}
\end{figure}

%% file: sections/04-experiment.tex
\section{Experiments}
\label{sec:experiments}
\input{floating/main_table}
\subsection{Experimental Setup}
\label{sec:exp_setup}
\paragraph{Dataset and Benchmarks.}
We evaluate \textsc{CamCue} on perspective-shift reasoning using \textsc{CamCue-Data}, and general multi-image spatial reasoning benchmarks, including MMSI~\citep{yang2025mmsi} and MindCube~\citep{yin2025spatial}, which do not provide explicit camera pose inputs. This allows us to verify that \textsc{CamCue} yields substantial gains on perspective-taking, while preserving general multi-image reasoning performance.

\paragraph{Backbones and baselines.}
We deploy \textsc{CamCue} on top of QwenVL2.5-3B, QwenVL2.5-7B~\citep{bai2025qwen2}, and InternVL2.5-8B~\citep{chen2024expanding}. We compare \textsc{CamCue} with the backbone-only setting and also MindJourney~\citep{yang2025mindjourney}, a competitive test-time scaling method that calls an external world model to generate auxiliary observations and feeds them back to the MLLM for answering. Although MindJourney is primarily formulated for a single-reference-view setting, it uses Stable Virtual Camera (SVC)~\citep{zhou2025stable} as the underlying world model, which can also condition on multiple contextual images with known camera poses to synthesize an imagined observation set at inference time. We therefore adapt MindJourney to our multi-view setting by providing the full set of contextual views and poses to SVC.

\paragraph{Training setup.}
We use mixed-data training by interleaving MindCube training examples without camera poses, mixing one MindCube example for every five \textsc{CamCue} training data. For such samples we omit pose loss and optimize only $\mathcal{L}_{\text{lang}}$, enabling the model to fall back to standard multi-image inference while still leveraging pose when provided. We fine-tune all backbones using LoRA with a cosine learning-rate scheduler and a batch size of 8. We train the pose-related modules with a learning rate of $5\times10^{-5}$ and the LoRA parameters with a learning rate of $1\times10^{-5}$, using a warmup ratio of $0.03$. More details are in the Appendix~\ref{app:training_hparams}.

\subsection{Experiment Results}
\paragraph{Main results on perspective-shift reasoning.}
Table~\ref{tab:persp_main} reports accuracy on \textsc{CamCue-Data}. \textsc{CamCue} yields consistent gains across all three backbones, with the largest improvements on viewpoint-sensitive categories such as visibility, distance order, and relative relation. Notably, MindJourney improves over backbone-only inference, confirming the benefit of inference-time imagination for perspective reasoning. However, it remains substantially below \textsc{CamCue}. A key reason is that MindJourney typically explores a set of navigational rollouts (e.g., turn left/right, move forward) to collect additional synthesized observations, rather than grounding the natural-language viewpoint description to a single explicit target viewpoint. As a result, the explored views may be informative but are not guaranteed to coincide with the queried perspective, motivating our explicit viewpoint grounding and pose-conditioned target-view synthesis.

\paragraph{Results on General Multi-image Benchmarks}
\input{floating/general_benchmark_table}
We further evaluate general multi-image spatial reasoning on MMSI and the MindCube Tiny benchmark.
Since these benchmarks do not provide camera pose information, we evaluate \textsc{CamCue} without target-view synthesis, where the model answers directly from the given contextual images.
As shown in Table~\ref{tab:mindcube_mmsi}, \textsc{CamCue} improves overall accuracy on both benchmarks, indicating that our training does not compromise general multi-image reasoning and can transfer beyond pose-supervised settings.

\label{sec:exp_ablation}

\paragraph{Camera Pose Prediction}

\input{floating/camera_pose_table}

Table~\ref{tab:pose_estimation} reports target pose prediction accuracy when the desired viewpoint is specified by a natural-language description. We consider two description sources: GPT-4.1 generated descriptions and human expert annotations. Following standard pose-evaluation practice, we report the fraction of samples whose rotation and translation errors fall within each threshold. Overall, \textsc{CamCue} achieves high target-pose prediction accuracy from natural language description. Under synthetic descriptions, 91.5\% of examples have a camera rotation error below 20$^\circ$, and 92.9\% have a translation error below 0.5; with human-written descriptions, these fractions increase to 100.0\% and 95.1\%, respectively. We attribute this gap to the fact that expert annotations are typically more detailed and less ambiguous than LLM-generated descriptions, providing clearer cues about the camera position. To qualitatively verify geometric fidelity, Figure~\ref{fig:render_example} visualizes the predicted pose by rendering the scene from the estimated camera and comparing it against the ground-truth target view. The close alignment in both viewpoint and visible content indicates that \textsc{CamCue} reliably grounds language to precise camera geometry, providing a dependable basis for downstream perspective-shift reasoning.

\subsection{Ablations and Analysis}
\label{sec:exp_ablation}

\paragraph{Ablation studies.}
\input{floating/ablation_camcue_table}

Table~\ref{tab:ablation_7b} analyzes Qwen2.5-VL-7B to disentangle the effects of pose supervision and imagined-view feedback.
QA-FT (fine-tuning with QA supervision only) yields only marginal changes from the base model, indicating that answer-only supervision does not reliably teach perspective shifting.
In contrast, Pose only model (training with pose supervision but no target-view synthesis at inference) already improves over QA-FT, showing that learning to predict and use camera pose provides a meaningful geometric prior even without imagined images.
Building on this, \textsc{CamCue} further introduces target-view synthesis and image feedback, leading to a substantial jump across all categories, which confirms the importance of converting viewpoint grounding into concrete visual evidence for answering.
Finally, \textsc{CamCue (GT)} replaces the imagined view with the ground-truth target view and serves as an oracle upper bound, suggesting additional headroom from improved novel-view synthesis quality.

\input{floating/render_example}
\paragraph{Faithful Viewpoint Imagination}
\input{floating/image_generation_table}

Camera pose is an effective intermediate variable for connecting multi-view contextual understanding with target-view inference. Predicting the target camera pose encourages the model to represent the viewpoint description in an explicit geometric form, which helps relate observations across contextual views under viewpoint change. This geometric anchor can then be used to guide a 3D-aware imagination step, where the synthesized target view is constrained by the predicted camera pose and is therefore better aligned with the ground-truth physical scene than pose-free generation that must infer geometry and viewpoint implicitly from images and language.

This distinction is critical when the imagined observation is used as evidence for answering. The generated target view must remain faithful and stick to the underlying physical environment, otherwise spurious details can mislead downstream reasoning. Despite being a strong generator, Nano Banana~\citep{team2023gemini} frequently hallucinates or drifts in layout, viewpoint, and object configurations, and can even perform worse than directly answering from the contextual views. As shown in Table~\ref{tab:nanobanana_baseline}, Nano Banana underperforms the Base setting by $4.33$\%. A stronger variant, Nano Banana Pro, produces more faithful images on average and yields some improvement, but its pose-free generations are still not reliably grounded and remain unstable in challenging cases. In contrast, \textsc{CamCue} uses the predicted camera pose as an explicit geometric anchor to constrain imagination, leading to a substantial gain and more dependable evidence for perspective-shift reasoning.

These failures are visible in Fig.~\ref{fig:render_example}: although Nano Banana often preserves the coarse semantics of the scene, its pose-free generations can drift from the ground-truth environment, such as altering global layout and background structures (Example 1–2), or keeping the scene content but misestimating the viewing direction and even changing object configuration (Example 3–4).

\vspace{-10pt}
\paragraph{Efficiency}
\input{floating/efficiency_table}
Table~\ref{tab:inference_time} compares inference-time cost per example. \textsc{CamCue} remains efficient in practice, it predicts the target pose in a single forward pass, and synthesizes the imagined observation with a feed-forward image decoder, enabling fast end-to-end feedback for interactive use. In contrast, MindJourney shows substantially higher latency because it performs test-time scaling via iterative search over multiple candidate rollouts and repeatedly queries the world model and VLM to aggregate evidence.

%% file: floating/main_table.tex
\begin{table*}[t]
\centering
\begingroup
\renewcommand{\arraystretch}{1.05}
\setlength{\tabcolsep}{6pt}
\caption{Main results on perspective-shift reasoning.}
\label{tab:persp_main}
\begin{tabular}{l c @{\hskip 10pt} c c c c c}
\toprule
\textbf{Model} & \textbf{Overall (Avg.)} & \textbf{Attribute} & \textbf{Visibility} & \textbf{Distance Order} & \textbf{Relative Relation} & \textbf{Count} \\
\cmidrule(lr){1-1}\cmidrule(lr){2-2}\cmidrule(lr){3-7}

\textbf{Qwen2.5-VL-7B}            & 71.06 & 93.00 & 84.31 & 71.43 & 59.73 & 55.29 \\
\qquad + MindJourney              & 72.83 & 92.00 & 84.31 & 80.22 & 65.75 & 50.59 \\
\rowcolor[HTML]{C9DAF8}
\rule{0pt}{1.1ex}\qquad + \textbf{CamCue} & \textbf{80.12} & 92.00 & 88.24 & 83.52 & 78.52 & 60.00 \\
\midrule

\textbf{Qwen2.5-VL-3B}            & 67.52 & 97.00 & 80.39 & 62.64 & 57.05 & 50.59 \\
\qquad + MindJourney              & 70.28 & 95.00 & 76.47 & 69.23 & 64.64 & 50.59 \\
\rowcolor[HTML]{C9DAF8}
\rule{0pt}{1.1ex}\qquad + \textbf{CamCue} & \textbf{75.92} & 94.12 & 82.35 & 80.43 & 67.97 & 63.53 \\
\midrule

\textbf{InternVL-2.5-8B}              & 68.11 & 89.00 & 76.47 & 80.22 & 58.56 & 52.94 \\
\qquad + MindJourney              & 74.21 & 93.00 & 82.35 & 85.71 & 64.64 & 55.29 \\
\rowcolor[HTML]{C9DAF8}
\rule{0pt}{1.1ex}\qquad + \textbf{CamCue} & \textbf{77.36} & 94.00 & 80.39 & 89.01 & 72.38 & 54.12 \\

\bottomrule
\end{tabular}
\endgroup

\end{table*}

%% file: floating/general_benchmark_table.tex
\newcolumntype{C}{>{\centering\arraybackslash}X}

\begin{table}[t]
\centering
\footnotesize
\setlength{\tabcolsep}{2.5pt}
\renewcommand{\arraystretch}{1.05}
\caption{Results on MindCube Tiny and MMSI with the Qwen2.5-VL-7B backbone.
Pose-Only denotes CamCue pose-only
inference without imagined image feedback.}
\resizebox{\columnwidth}{!}{%
\begin{tabularx}{\columnwidth}{l C C C C}
\toprule
\multicolumn{1}{l}{\textbf{Model}} &
\multicolumn{4}{c}{\textbf{MindCube Tiny Accuracy (\%)}}\\
\cmidrule(lr){2-5}
& \textbf{Overall} & \textbf{Rotation} & \textbf{Among} & \textbf{Around}\\
\midrule
Base & 29.3 & 38.7 & 29.5 & 21.4 \\
Pose-Only & \textbf{47.43} & 75.00 & 31.67 & 63.20 \\
\bottomrule
\end{tabularx}%
}

\vspace{6pt}

\resizebox{\columnwidth}{!}{%
\begin{tabularx}{\columnwidth}{l C C C C C}
\toprule
\multicolumn{1}{l}{\textbf{Model}} &
\multicolumn{5}{c}{\textbf{MMSI Accuracy (\%)}}\\
\cmidrule(lr){2-6}
& \textbf{Overall} & \textbf{Position} & \textbf{Attribute} & \textbf{Motion} & \textbf{MSR}\\
\midrule
Base & 25.9 & 25.9 & 21.5 & 30.0 & 25.8 \\
Pose-Only & \textbf{28.8} & 29.9 & 29.2 & 27.3 & 26.8 \\
\bottomrule
\end{tabularx}%
}

\label{tab:mindcube_mmsi}
\end{table}

%% file: floating/camera_pose_table.tex
\begin{table}[t]
\centering
\small
\setlength{\tabcolsep}{5pt}
\renewcommand{\arraystretch}{1.05}
\caption{Camera pose estimation accuracy under different viewpoint description sources.
Values are the percentage of samples with rotation/translation error within each threshold.}
\resizebox{\columnwidth}{!}{%
\begin{tabular}{lccc|ccc}
\toprule
\textbf{Viewpoint Description} &
\multicolumn{3}{c|}{\textbf{Rotation Acc.} $\uparrow$ (\%)} &
\multicolumn{3}{c}{\textbf{Translation Acc.} $\uparrow$ (\%)} \\
\cmidrule(lr){2-4}\cmidrule(lr){5-7}
& R@5$^\circ$ & R@10$^\circ$ & R@20$^\circ$ & t@0.1 & t@0.3 & t@0.5 \\
\midrule
Synthetic          & 19.3 & 35.4 & 91.5  & 12.0 & 62.4 & 92.9 \\
Human Description  & 30.1 & 56.9 & 100.0 & 19.5 & 74.8 & 95.1 \\
\bottomrule
\end{tabular}%
}
\label{tab:pose_estimation}
\end{table}

%% file: floating/ablation_camcue_table.tex
\begin{table}[t]
\centering
\footnotesize
\setlength{\tabcolsep}{4pt}
\renewcommand{\arraystretch}{1.02}
\caption{Ablation study on Qwen2.5-VL-7B backbone.
(1) fine-tuning with QA supervision only (no pose).
(2) pose-only inference without imagined image feedback.
(3) full pipeline with imagined image feedback.
(4) oracle upper bound by replacing the imagined image with the GT target view.}
\resizebox{\linewidth}{!}{%
\begin{tabular}{c lcccccc}
\toprule
 & \textbf{Method} & \textbf{Avg.} & \textbf{Attr.} & \textbf{Vis.} & \textbf{Dist.} & \textbf{Rel.} & \textbf{Cnt.} \\
\midrule
(0) & Base         & 71.06 & 93.00 & 84.31 & 71.43 & 59.73 & 55.29 \\
(1) & QA-FT        & 71.26 & 92.00 & 78.43 & 67.03 & 65.75 & 58.82 \\
(2) & Pose-Only & 72.44 & 90.00 & 80.39 & 80.22 & 65.19 & 54.12 \\
(3) & CamCue       & 80.12 & 92.00 & 88.24 & 83.52 & 78.52 & 60.00 \\
(4) & CamCue (GT)  & \textbf{87.20} & 98.00 & 98.04 & 90.11 & 83.43 & 72.94 \\
\bottomrule
\end{tabular}%
}
\label{tab:ablation_7b}
\end{table}

%% file: floating/render_example.tex
\begin{figure*}[!t]
    \centering
    \includegraphics[width=0.9\textwidth]{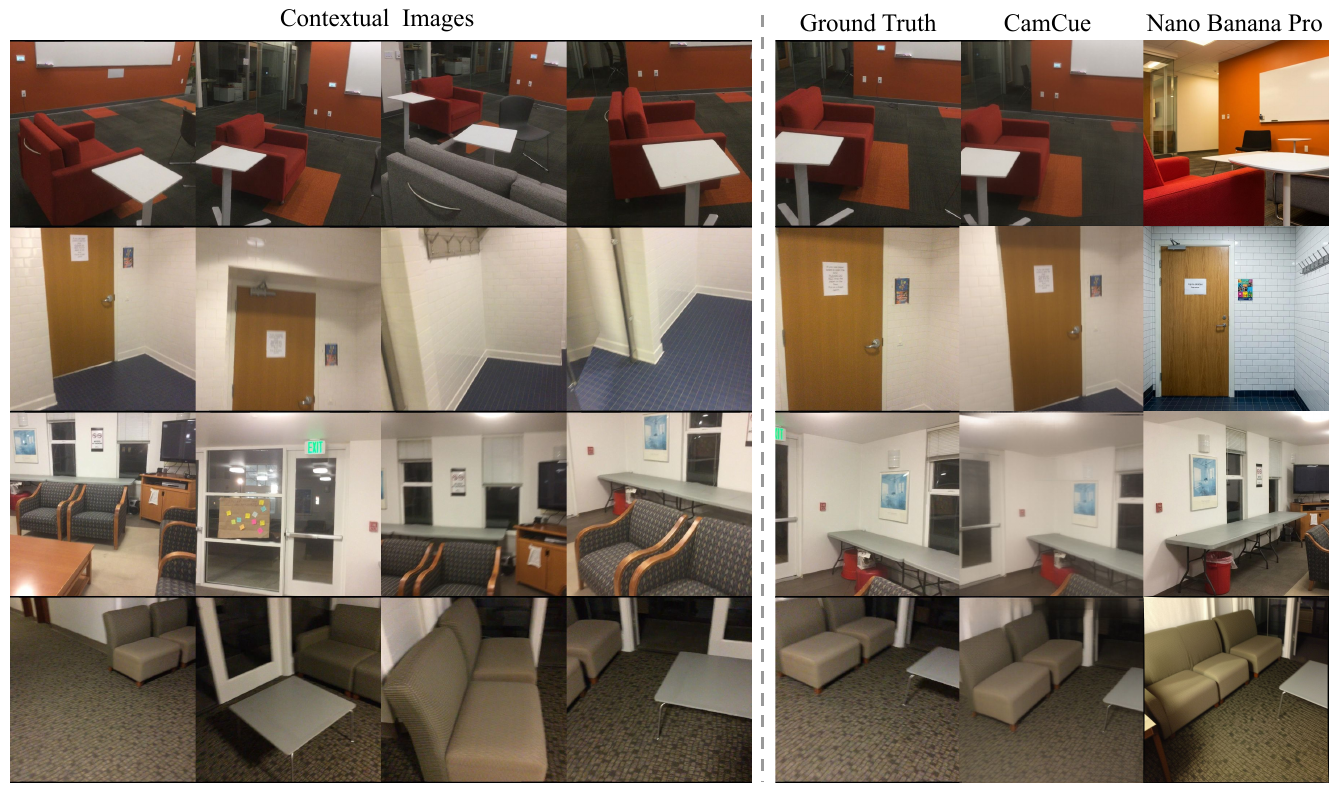}
    \caption{Qualitative comparison of imagined target views.}
    \label{fig:render_example}
\end{figure*}

%% file: floating/image_generation_table.tex
\newcolumntype{C}{>{\centering\arraybackslash}X}

\begin{table}[t]
\centering
\small
\setlength{\tabcolsep}{2.8pt}
\renewcommand{\arraystretch}{1.05}
\caption{Comparison with a pose-free image-generation baseline.
All methods use the same 7B MLLM backbone for answering (Qwen2.5-VL-7B). Nano Banana / Nano Banana Pro synthesize the target view from multi-view context images and a viewpoint description, and the generated image is fed back to the backbone for QA.}
\label{tab:nanobanana_baseline}
\begin{tabularx}{\columnwidth}{l C C C C C C}
\toprule
\textbf{Method} & \textbf{Avg.} & \textbf{Attr.} & \textbf{Vis.} & \textbf{Dist.} & \textbf{Rel.} & \textbf{Cnt.} \\
\midrule
Base            & 71.06 & 93.00 & 84.31 & 71.43 & 59.73 & 55.29 \\
Nano Banana     & 66.73 & 82.00 & 76.47 & 75.82 & 58.56 & 50.59 \\
Nano Banana Pro & 76.77 & 95.00 & 94.12 & 82.42 & 69.61 & 54.12 \\
CamCue          & \textbf{80.12} & 92.00 & 88.24 & 83.52 & 78.52 & 60.00 \\
\bottomrule
\end{tabularx}
\end{table}

%% file: floating/efficiency_table.tex
\begin{table}[t]
\centering
\small
\setlength{\tabcolsep}{6pt}
\renewcommand{\arraystretch}{1.05}
\caption{Inference-time cost per example.}
\label{tab:inference_time}

\begin{tabular}{@{}lccc@{}}
\toprule
\textbf{Method}  & \textbf{CamCue} & \textbf{Nano Banana} & \textbf{MindJourney} \\
\midrule
Time (s)  & 1.453 & 35.1 & 256.6 \\
\bottomrule
\end{tabular}%
\end{table}

%% file: sections/05-discussion.tex
\section{Limitations}
Our study focuses on perspective-shift question answering, which provides a clean setting to evaluate viewpoint grounding, but does not directly cover embodied planning or action. In the bigger picture, pose-grounded viewpoint imagination could serve as an additional evidence source in embodied pipelines, where language-specified viewpoints guide what to seek or simulate beyond the observed context. However, when the synthesized imagination is noisy or visually ambiguous, it may hurt reasoning, especially for small objects or fine-grained spatial relations. Developing a reliability-aware strategy that estimates when the imagined view is informative and selectively uses it. Otherwise, the model should fall back to the original method over the original observations. This is an important direction for future work.

%% file: sections/06-conclusion.tex
\section{Conclusion}

We propose \textsc{CamCue}, a pose-aware framework that equips MLLMs with explicit viewpoint grounding for multi-image spatial reasoning. \textsc{CamCue} predicts the target camera pose from multi-view observations and a natural language viewpoint description, and uses this pose as an anchor to support target-view inference. Building on the predicted camera pose, our pipeline can optionally synthesize a target-view observation as additional evidence, providing a more faithful form of viewpoint imagination for downstream reasoning tasks. Across experiments, \textsc{CamCue} consistently improves overall performance over strong baselines, while remaining efficient in practice, substantially reducing inference-time cost compared to prior methods that rely on iterative test-time search.

%% file: sections/07-acknowledgementy.tex
\section{Acknowledgments}
This research is based upon work supported by U.S. DARPA ECOLE Program No. \#HR00112390060. The views and conclusions contained herein are those of the authors and should not be interpreted as necessarily representing the official policies, either expressed or implied, of DARPA, or the U.S. Government. The U.S. Government is authorized to reproduce and distribute reprints for governmental purposes notwithstanding any copyright annotation therein.

%% file: sections/a_appendix.tex
\newpage
\appendix
\onecolumn

\section*{Appendix}
\appendix
\section{Data Curation}
\subsection{View Selection Details}
\label{app:view_selection}

\input{floating/data_alg}
\input{floating/depth_alg}
\paragraph{Parameters.}
For candidate context filtering, we keep frames within a moderate translation range to the target,
$d\in(0.4, 2.5)$ meters, and avoid near-duplicate views using a distinctness heuristic:
$d>0.6$ m, or $\Delta\theta>15^\circ$.
We then greedily select 4 context views to maximize depth-based visibility coverage (Algorithm~\ref{alg:visibility}),
and keep a target--context group only if the overall coverage satisfies $|M|/|P|\ge\gamma$ with $\gamma=0.80$.
Finally, we remove redundant groups by discarding a candidate target view if it is too similar to an existing one,
using a translation threshold $\tau_t=0.5$ m and a rotation threshold $\tau_\theta=45^\circ$.

\input{floating/data_samples}

\subsection{Data Samples}
\label{app:data_samples}
Figure~\ref{fig:data_samples} shows three representative examples from \textsc{CamCue-Data}. Each datapoint consists of four contextual images, a natural-language description specifying a target viewpoint, the target-view image, and QA pairs that must be answered from the described target perspective.

\paragraph{Example 1 (Kitchen scene).}

\textbf{Target-view description:} \emph{The camera is to the right of the stove and counter, near the corner where the counter meets the wall. It is facing the wall with windows, with the stove and counter on the left side of the view and the heater on the right side of the view.}

\textbf{Count:} \emph{How many windows are visible in this image?}

\textbf{Options:} (A) 1 \;\; (B) 2 \;\; (C) 3 \;\; (D) 4. \;\;
\textbf{Answer:} (B).

\textbf{Visibility:} \emph{Can you see the fire extinguisher from this viewpoint?}

\textbf{Options:} (A) Yes \;\; (B) No. \;\;
\textbf{Answer:} (A).

\paragraph{Example 2 (Auditorium).}
\textbf{Target-view description:} \emph{The camera is to the right of the rows of auditorium chairs, close to the wall with the handrail. It is aimed toward the wall with the handrail and door, with the chairs on the left side of the view and the wall on the right side of the view.}

\textbf{Relative relation:} \emph{Where is the handrail located relative to the seats in this image?}

\textbf{Options:} (A) Behind \;\; (B) In front of \;\; (C) Left of \;\; (D) Right of. \;\;
\textbf{Answer:} (A).

\textbf{Attribute:} \emph{What is the most likely material of the handrail visible on the wall?}

\textbf{Options:} (A) Wood \;\; (B) Metal \;\; (C) Plastic \;\; (D) Glass. \;\;
\textbf{Answer:} (A).

\paragraph{Example 3 (Gym).}
\textbf{Target-view description:} \emph{The camera is placed to the right of the treadmills, near where the mat meets the wooden floor. It is facing toward the dumbbell rack on the left side of the view and the curved wall on the right side of the view.}

\textbf{Distance order:} \emph{Which is closer to you, the treadmill or the dumbbell rack?}

\textbf{Options:} (A) The treadmill \;\; (B) The dumbbell rack. \;\;
\textbf{Answer:} (A).

\section{Training Hyperparameters}
\label{app:training_hparams}
\vspace{0.5em}
\begin{table}[h]
\centering
\small
\setlength{\tabcolsep}{5pt}
\caption{Training hyperparameters.}
\label{tab:opt_hparams}
\begin{tabular}{lcc}
\toprule
\textbf{Hyperparameter} & \textbf{Qwen2.5-VL (3B/7B)} & \textbf{InternVL2.5 (8B)} \\
\midrule
\#GPUs & 4 & 4 \\
Per-device batch size & 2 & 2 \\
Effective batch size & 8 & 8  \\
Training steps & 9000 & 9000 \\
Scheduler & Cosine & Cosine \\
Warmup ratio & 0.03 & 0.03 \\
\midrule
Geometry LR & $5\times10^{-5}$ & $5\times10^{-5}$ \\
Language (LoRA) LR & $1\times10^{-5}$ & $1\times10^{-5}$ \\
\bottomrule
\end{tabular}

\end{table}

\begin{table}[h]
\centering
\small
\setlength{\tabcolsep}{5pt}
\caption{Model adaptation and loss weights.}
\label{tab:model_loss_hparams}
\begin{tabular}{lcc}
\toprule
\textbf{Setting} & \textbf{Qwen2.5-VL (3B/7B)} & \textbf{InternVL2.5 (8B)} \\
\midrule
LoRA rank $r$ & 16 & 16 \\
LoRA $\alpha$ & 64 & 64 \\
LoRA dropout & 0.10 & 0.10 \\
LoRA target modules & \multicolumn{2}{c}{\{q,k,v,o proj, gate/up/down proj\}} \\
\midrule
Pose loss weight $\lambda_{\text{pose}}$ & 0.2 & 0.2 \\
Language loss weight $\lambda_{\text{lang}}$ & 1.0 & 1.0 
\\
\bottomrule
\end{tabular}

\end{table}

Tables~\ref{tab:opt_hparams}--\ref{tab:model_loss_hparams} summarize the key training hyperparameters used in our experiments. We fine-tune the language backbone with LoRA while jointly training the pose-related modules.
The overall objective is a weighted sum of language modeling loss and pose regression loss with $\lambda_{\text{lang}}=1.0$ and $\lambda_{\text{pose}}=0.2$.

\section{Additional qualitative examples and failure cases}

\input{floating/failure_case}
In Figure~\ref{fig:failure_case}, we compare the ground-truth target view with synthesized views from \textsc{CamCue} and Nano Banana. \textsc{CamCue} generally preserves the scene layout but may produce blurred renderings in some cases. In contrast, Nano Banana often generates visually sharp images but may exhibit inaccurate viewpoint estimation, and can introduce changes to the environment (e.g., modifying object placements or adding/removing scene elements), which breaks geometric and physical consistency for spatial reasoning.

%% file: floating/data_alg.tex
\begin{algorithm}[H]
\caption{Multi-view group selection with depth-based visibility.}
\label{alg:view_selection}
\begin{algorithmic}[1]
\State $\mathcal{G}\gets\emptyset$
\For{each target view $t$}
  \State $\mathcal{C}\gets\{c\neq t \mid \mathrm{PoseFilter}(t,c)\}$
  \If{$|\mathcal{C}|<4$} \State \textbf{continue} \EndIf
  \State $P\gets \mathrm{TargetSamples}(t)$
  \State $\mathcal{S}\gets\emptyset$;\; $M\gets\emptyset$
  \For{$r=1$ to $4$}
    \State Choose $c^\star\in\mathcal{C}\setminus\mathcal{S}$ maximizing $|\mathrm{Vis}(P,t,c)\setminus M|$
    \State $\mathcal{S}\gets\mathcal{S}\cup\{c^\star\}$;\; $M\gets M\cup \mathrm{Vis}(P,t,c^\star)$
  \EndFor
  \If{$|M|/|P|<\gamma$} \State \textbf{continue} \EndIf
  \If{$\mathrm{Redundant}(t,\mathcal{S},\mathcal{G})$} \State \textbf{continue} \EndIf
  \State $\mathcal{G}\gets\mathcal{G}\cup\{(t,\mathcal{S})\}$
\EndFor
\State \Return $\mathcal{G}$
\end{algorithmic}
\end{algorithm}

%% file: floating/depth_alg.tex
\begin{algorithm}[H]
\caption{Depth-based visibility test used in $\mathrm{Vis}(P,t,c)$.}
\label{alg:visibility}
\begin{algorithmic}[1]
\State \textbf{Procedure} $\mathrm{Vis}(P,t,c)$
\State $V\gets\emptyset$
\For{each sample $p\in P$}
  \State Back-project $p$ in target view using $(D_t,K,C_t)$ to obtain a 3D point $X$
  \State Project $X$ into context view $c$ to get $(u,v,\hat{z})$
  \If{$(u,v)$ is in bounds \textbf{and} $\hat{z}\le D_c(u,v)+\epsilon$}
    \State $V\gets V\cup\{p\}$
  \EndIf
\EndFor
\State \Return $V$
\end{algorithmic}
\end{algorithm}

%% file: floating/data_samples.tex
\begin{figure*}[!t]
    \centering
    \includegraphics[width=0.9\textwidth]{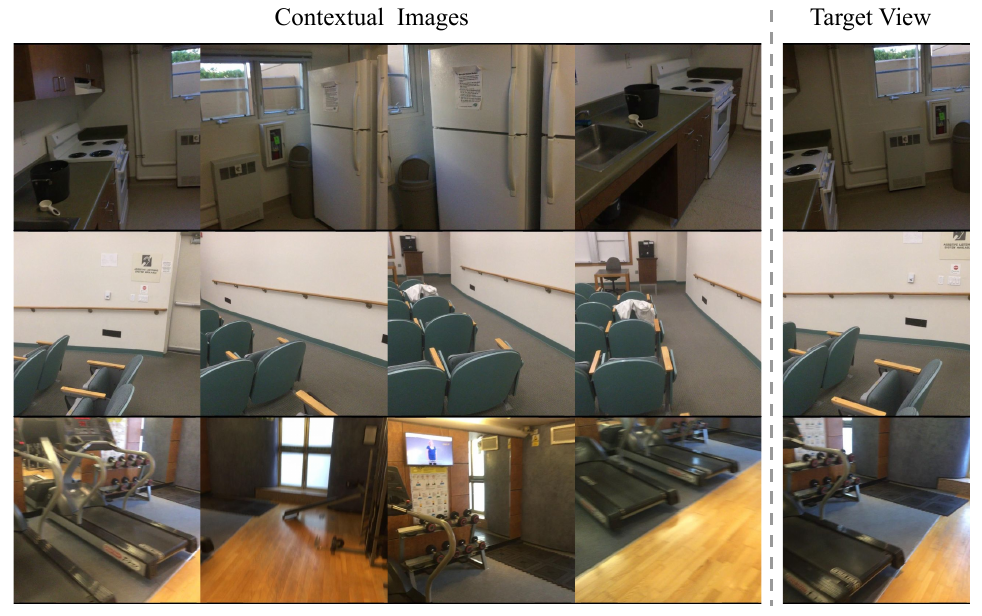}

    \caption{Data Samples from CamCue}
    \label{fig:data_samples}
\end{figure*}

%% file: floating/failure_case.tex
\begin{figure*}[!t]
    \centering
    \includegraphics[width=0.6\textwidth]{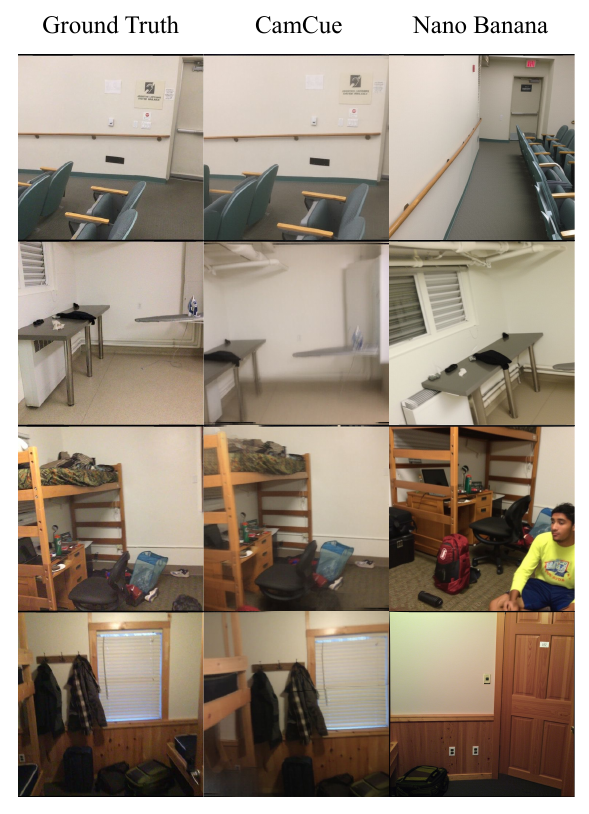}

    \caption{Additional qualitative examples and failure cases.}
    \label{fig:failure_case}
\end{figure*}

%% file: example_paper.bib
@inproceedings{chen2024and,
  title={Where am i and what will i see: An auto-regressive model for spatial localization and view prediction},
  author={Chen, Junyi and Huang, Di and Ye, Weicai and Ouyang, Wanli and He, Tong},
  booktitle={The Thirteenth International Conference on Learning Representations},
  year={2024}
}

@article{yeh2025seeinganotherperspectiveevaluating,
  title={Seeing from Another Perspective: Evaluating Multi-View Understanding in MLLMs},
  author={Chun-Hsiao Yeh and Chenyu Wang and Shengbang Tong and Ta-Ying Cheng and Rouyu Wang and Tianzhe Chu and Yuexiang Zhai and Yubei Chen and Shenghua Gao and Yi Ma},
  journal={ArXiv},
  year={2025},
  volume={abs/2504.15280},
  url={https://api.semanticscholar.org/CorpusID:277955366}
}

@article{liu2025spatialreasoningmultimodallarge,
  title={Spatial Reasoning in Multimodal Large Language Models: A Survey of Tasks, Benchmarks and Methods},
  author={Weichen Liu and Qiyao Xue and Haoming Wang and Xiangyu Yin and Boyuan Yang and Wei Gao},
  journal={ArXiv},
  year={2025},
  volume={abs/2511.15722},
  url={https://api.semanticscholar.org/CorpusID:283109806}
}

@misc{zhang2025mllmsstrugglespatialunderstanding,
      title={Why Do MLLMs Struggle with Spatial Understanding? A Systematic Analysis from Data to Architecture}, 
      author={Wanyue Zhang and Yibin Huang and Yangbin Xu and JingJing Huang and Helu Zhi and Shuo Ren and Wang Xu and Jiajun Zhang},
      year={2025},
      eprint={2509.02359},
      archivePrefix={arXiv},
      primaryClass={cs.CV},
      url={https://arxiv.org/abs/2509.02359}, 
}

@misc{li2025viewspatialbenchevaluatingmultiperspectivespatial,
      title={ViewSpatial-Bench: Evaluating Multi-perspective Spatial Localization in Vision-Language Models},
      author={Dingming Li and Hongxing Li and Zixuan Wang and Yuchen Yan and Hang Zhang and Siqi Chen and Guiyang Hou and Shengpei Jiang and Wenqi Zhang and Yongliang Shen and Weiming Lu and Yueting Zhuang},
      year={2025},
      eprint={2505.21500},
      archivePrefix={arXiv},
      primaryClass={cs.CV},
      url={https://arxiv.org/abs/2505.21500},
}

@article{yu2025farvlmsvisualspatial,
  title={How Far are VLMs from Visual Spatial Intelligence? A Benchmark-Driven Perspective},
  author={Songsong Yu and Yuxin Chen and Hao Ju and Lianjie Jia and Fuxi Zhang and Shaofei Huang and Yuhan Wu and Rundi Cui and Binghao Ran and Zaibin Zhang and Zhedong Zheng and Zhipeng Zhang and Yifan Wang and Lin Song and Lijun Wang and Yanwei Li and Ying Shan and Huchuan Lu},
  journal={ArXiv},
  year={2025},
  volume={abs/2509.18905},
  url={https://api.semanticscholar.org/CorpusID:281496332}
}

@misc{sqa3d,
      title={SQA3D: Situated Question Answering in 3D Scenes}, 
      author={Xiaojian Ma and Silong Yong and Zilong Zheng and Qing Li and Yitao Liang and Song-Chun Zhu and Siyuan Huang},
      year={2023},
      eprint={2210.07474},
      archivePrefix={arXiv},
      primaryClass={cs.CV},
      url={https://arxiv.org/abs/2210.07474}, 
}

@article{gholami2025spatial,
  title={Spatial reasoning with vision-language models in ego-centric multi-view scenes},
  author={Gholami, Mohsen and Rezaei, Ahmad and Weimin, Zhou and Mao, Sitong and Zhou, Shunbo and Zhang, Yong and Akbari, Mohammad},
  journal={arXiv preprint arXiv:2509.06266},
  year={2025}
}

@article{wang20253d,
  title={3D Question Answering via only 2D Vision-Language Models},
  author={Wang, Fengyun and Yu, Sicheng and Wu, Jiawei and Tang, Jinhui and Zhang, Hanwang and Sun, Qianru},
  journal={arXiv preprint arXiv:2505.22143},
  year={2025}
}

@inproceedings{yin2025spatial,
  title={Spatial mental modeling from limited views},
  author={Yin, Baiqiao and Wang, Qineng and Zhang, Pingyue and Zhang, Jianshu and Wang, Kangrui and Wang, Zihan and Zhang, Jieyu and Chandrasegaran, Keshigeyan and Liu, Han and Krishna, Ranjay and others},
  booktitle={Structural Priors for Vision Workshop at ICCV'25},
  year={2025}
}

@article{zhao2025spacemind,
  title={SpaceMind: Camera-Guided Modality Fusion for Spatial Reasoning in Vision-Language Models},
  author={Zhao, Ruosen and Zhang, Zhikang and Xu, Jialei and Chang, Jiahao and Chen, Dong and Li, Lingyun and Sun, Weijian and Wei, Zizhuang},
  journal={arXiv preprint arXiv:2511.23075},
  year={2025}
}

@article{lee2025perspective,
  title={Perspective-aware reasoning in vision-language models via mental imagery simulation},
  author={Lee, Phillip Y and Je, Jihyeon and Park, Chanho and Uy, Mikaela Angelina and Guibas, Leonidas and Sung, Minhyuk},
  journal={arXiv preprint arXiv:2504.17207},
  year={2025}
}

@article{yang2025mmsi,
  title={MMSI-Bench: A Benchmark for Multi-Image Spatial Intelligence},
  author={Yang, Sihan and Xu, Runsen and Xie, Yiman and Yang, Sizhe and Li, Mo and Lin, Jingli and Zhu, Chenming and Chen, Xiaochen and Duan, Haodong and Yue, Xiangyu and others},
  journal={arXiv preprint arXiv:2505.23764},
  year={2025}
}

@article{article,
author = {Wang, Ranxiao},
year = {2012},
month = {05},
pages = {575-87},
title = {Theories of spatial representations and reference frames: What can configuration errors tell us?},
volume = {19},
journal = {Psychonomic bulletin \& review},
doi = {10.3758/s13423-012-0258-2}
}

@article{Meilinger2011,
  author  = {Meilinger, Tobias and Berthoz, Alain and Wiener, Jan M.},
  title   = {The integration of spatial information across different viewpoints},
  journal = {Memory \& Cognition},
  year    = {2011},
  volume  = {39},
  number  = {6},
  pages   = {1042--1054},
  doi     = {10.3758/s13421-011-0088-x},
  issn    = {1532-5946},
  url     = {https://doi.org/10.3758/s13421-011-0088-x}
}

@article{xu2025spatialbench,
  title={SpatialBench: Benchmarking Multimodal Large Language Models for Spatial Cognition},
  author={Xu, Peiran and Wang, Sudong and Zhu, Yao and Li, Jianing and Zhang, Yunjian},
  journal={arXiv preprint arXiv:2511.21471},
  year={2025}
}

@article{yang2025mindjourney,
  title={MindJourney: Test-Time Scaling with World Models for Spatial Reasoning},
  author={Yang, Yuncong and Liu, Jiageng and Zhang, Zheyuan and Zhou, Siyuan and Tan, Reuben and Yang, Jianwei and Du, Yilun and Gan, Chuang},
  journal={arXiv preprint arXiv:2507.12508},
  year={2025}
}

@article{jin2024lvsm,
  title={Lvsm: A large view synthesis model with minimal 3d inductive bias},
  author={Jin, Haian and Jiang, Hanwen and Tan, Hao and Zhang, Kai and Bi, Sai and Zhang, Tianyuan and Luan, Fujun and Snavely, Noah and Xu, Zexiang},
  journal={arXiv preprint arXiv:2410.17242},
  year={2024}
}

@article{zhou2025stable,
  title={Stable virtual camera: Generative view synthesis with diffusion models},
  author={Zhou, Jensen and Gao, Hang and Voleti, Vikram and Vasishta, Aaryaman and Yao, Chun-Han and Boss, Mark and Torr, Philip and Rupprecht, Christian and Jampani, Varun},
  journal={arXiv preprint arXiv:2503.14489},
  year={2025}
}

@article{liao2025thinking,
  title={Thinking with Camera: A Unified Multimodal Model for Camera-Centric Understanding and Generation},
  author={Liao, Kang and Wu, Size and Wu, Zhonghua and Jin, Linyi and Wang, Chao and Wang, Yikai and Wang, Fei and Li, Wei and Loy, Chen Change},
  journal={arXiv preprint arXiv:2510.08673},
  year={2025}
}

@inproceedings{yang2025thinking,
  title={Thinking in space: How multimodal large language models see, remember, and recall spaces},
  author={Yang, Jihan and Yang, Shusheng and Gupta, Anjali W and Han, Rilyn and Fei-Fei, Li and Xie, Saining},
  booktitle={Proceedings of the Computer Vision and Pattern Recognition Conference},
  pages={10632--10643},
  year={2025}
}

@inproceedings{yu2021pixelnerf,
  title={pixelnerf: Neural radiance fields from one or few images},
  author={Yu, Alex and Ye, Vickie and Tancik, Matthew and Kanazawa, Angjoo},
  booktitle={Proceedings of the IEEE/CVF conference on computer vision and pattern recognition},
  pages={4578--4587},
  year={2021}
}

@inproceedings{charatan2024pixelsplat,
  title={pixelsplat: 3d gaussian splats from image pairs for scalable generalizable 3d reconstruction},
  author={Charatan, David and Li, Sizhe Lester and Tagliasacchi, Andrea and Sitzmann, Vincent},
  booktitle={Proceedings of the IEEE/CVF conference on computer vision and pattern recognition},
  pages={19457--19467},
  year={2024}
}

@inproceedings{chen2024mvsplat,
  title={Mvsplat: Efficient 3d gaussian splatting from sparse multi-view images},
  author={Chen, Yuedong and Xu, Haofei and Zheng, Chuanxia and Zhuang, Bohan and Pollefeys, Marc and Geiger, Andreas and Cham, Tat-Jen and Cai, Jianfei},
  booktitle={European Conference on Computer Vision},
  pages={370--386},
  year={2024},
  organization={Springer}
}

@inproceedings{wu2024reconfusion,
  title={Reconfusion: 3d reconstruction with diffusion priors},
  author={Wu, Rundi and Mildenhall, Ben and Henzler, Philipp and Park, Keunhong and Gao, Ruiqi and Watson, Daniel and Srinivasan, Pratul P and Verbin, Dor and Barron, Jonathan T and Poole, Ben and others},
  booktitle={Proceedings of the IEEE/CVF conference on computer vision and pattern recognition},
  pages={21551--21561},
  year={2024}
}

@article{gao2024cat3d,
  title={Cat3d: Create anything in 3d with multi-view diffusion models},
  author={Gao, Ruiqi and Holynski, Aleksander and Henzler, Philipp and Brussee, Arthur and Martin-Brualla, Ricardo and Srinivasan, Pratul and Barron, Jonathan T and Poole, Ben},
  journal={arXiv preprint arXiv:2405.10314},
  year={2024}
}

@article{bai2025qwen2,
  title={Qwen2. 5-vl technical report},
  author={Bai, Shuai and Chen, Keqin and Liu, Xuejing and Wang, Jialin and Ge, Wenbin and Song, Sibo and Dang, Kai and Wang, Peng and Wang, Shijie and Tang, Jun and others},
  journal={arXiv preprint arXiv:2502.13923},
  year={2025}
}

@article{chen2024expanding,
  title={Expanding performance boundaries of open-source multimodal models with model, data, and test-time scaling},
  author={Chen, Zhe and Wang, Weiyun and Cao, Yue and Liu, Yangzhou and Gao, Zhangwei and Cui, Erfei and Zhu, Jinguo and Ye, Shenglong and Tian, Hao and Liu, Zhaoyang and others},
  journal={arXiv preprint arXiv:2412.05271},
  year={2024}
}

@article{cao2025spatialdreamer,
  title={SpatialDreamer: Incentivizing Spatial Reasoning via Active Mental Imagery},
  author={Cao, Meng and Li, Xingyu and Liu, Xue and Reid, Ian and Liang, Xiaodan},
  journal={arXiv preprint arXiv:2512.07733},
  year={2025}
}

@inproceedings{azuma2022scanqa,
  title={Scanqa: 3d question answering for spatial scene understanding},
  author={Azuma, Daichi and Miyanishi, Taiki and Kurita, Shuhei and Kawanabe, Motoaki},
  booktitle={proceedings of the IEEE/CVF conference on computer vision and pattern recognition},
  pages={19129--19139},
  year={2022}
}

@article{ray2024sat,
  title={SAT: Dynamic Spatial Aptitude Training for Multimodal Language Models},
  author={Ray, Arijit and Duan, Jiafei and Brown, Ellis and Tan, Reuben and Bashkirova, Dina and Hendrix, Rose and Ehsani, Kiana and Kembhavi, Aniruddha and Plummer, Bryan A and Krishna, Ranjay and others},
  journal={arXiv preprint arXiv:2412.07755},
  year={2024}
}

@article{jiang2025rayzer,
  title={RayZer: A Self-supervised Large View Synthesis Model},
  author={Jiang, Hanwen and Tan, Hao and Wang, Peng and Jin, Haian and Zhao, Yue and Bi, Sai and Zhang, Kai and Luan, Fujun and Sunkavalli, Kalyan and Huang, Qixing and others},
  journal={arXiv preprint arXiv:2505.00702},
  year={2025}
}

@article{team2023gemini,
  title={Gemini: a family of highly capable multimodal models},
  author={Team, Gemini and Anil, Rohan and Borgeaud, Sebastian and Alayrac, Jean-Baptiste and Yu, Jiahui and Soricut, Radu and Schalkwyk, Johan and Dai, Andrew M and Hauth, Anja and Millican, Katie and others},
  journal={arXiv preprint arXiv:2312.11805},
  year={2023}
}

@article{wu2025qwen,
  title={Qwen-image technical report},
  author={Wu, Chenfei and Li, Jiahao and Zhou, Jingren and Lin, Junyang and Gao, Kaiyuan and Yan, Kun and Yin, Sheng-ming and Bai, Shuai and Xu, Xiao and Chen, Yilei and others},
  journal={arXiv preprint arXiv:2508.02324},
  year={2025}
}

@article{achiam2023gpt,
  title={Gpt-4 technical report},
  author={Achiam, Josh and Adler, Steven and Agarwal, Sandhini and Ahmad, Lama and Akkaya, Ilge and Aleman, Florencia Leoni and Almeida, Diogo and Altenschmidt, Janko and Altman, Sam and Anadkat, Shyamal and others},
  journal={arXiv preprint arXiv:2303.08774},
  year={2023}
}
